\documentclass[letterpaper, 10 pt, conference]{ieeeconf}  

\IEEEoverridecommandlockouts                              

\UseRawInputEncoding
\usepackage{cite}
\usepackage{amsmath,amssymb,amsfonts}
\usepackage{algorithmic}
\usepackage{graphicx}
\usepackage{textcomp}
\usepackage{xcolor}
\usepackage{booktabs}
\usepackage{hyperref}
\usepackage{booktabs}
\usepackage{float}
\usepackage{subfig}

\title{\LARGE \bf
Fast2comm:~Collaborative perception combined with prior knowledge
}

\author{Zhengbin Zhang$^{1}$, Yan Wu$^{1,*}$, and Hongkun Zhang$^1$  
\thanks{$^{1}$School of Computer Science and Technology, Tongji University, Shanghai, China. ({\tt\small 2410958@tongji.edu.cn; yanwu@tongji.edu.cn; 2432259@tongji.edu.cn})}%
\thanks{*Corresponding author: Yan Wu}
}

\begin{document}

\maketitle
\thispagestyle{empty}
\pagestyle{empty}

\begin{abstract}
	Collaborative perception has the potential to significantly enhance perceptual accuracy through the sharing of complementary information among agents. However, real-world collaborative perception faces persistent challenges, particularly in balancing perception performance and bandwidth limitations, as well as coping with localization errors. To address these challenges, we propose \textit{Fast2comm}, a prior knowledge-based collaborative perception framework. Specifically, (1)~we propose a prior-supervised confidence feature generation method, that effectively distinguishes foreground from background by producing highly discriminative confidence features; (2)~we propose GT Bounding Box-based spatial prior feature selection strategy to ensure that only the most informative prior-knowledge features are selected and shared, thereby minimizing background noise and optimizing bandwidth efficiency while enhancing adaptability to localization inaccuracies; (3)~we decouple the feature fusion strategies between model training and testing phases, enabling dynamic bandwidth adaptation. To comprehensively validate our framework, we conduct extensive experiments on both real-world and simulated datasets. The results demonstrate the superior performance of our model and highlight the necessity of the proposed methods. Our code is available at \href{https://github.com/Zhangzhengbin-TJ/Fast2comm}{https://github.com/Zhangzhengbin-TJ/Fast2comm}.
\end{abstract}

\section{Introduction}
Single-agent or single-vehicle perception inevitably suffers from limitations such as occlusion and reduced long-distance detection capability. Recently, the advent of collaborative perception technologies~\cite{v2vnet,v2xvit,earlyfusion} has significantly advanced vehicle perception by enabling agents to share supplementary perceptual information, thereby facilitating more comprehensive and holistic perception. Such methods are crucial across a wide range of practical applications, including vehicle-to-everything (V2X) autonomous driving~\cite{v2vnet,disconet} and multi-robot warehouse automation systems~\cite{warehouse}.

However, in real-world scenarios, cooperative perception systems often struggle to provide sufficient real-time bandwidth, particularly when sharing raw data or a large volume of features. Moreover, GPS localization noise and asynchronous sensor measurements across agents can introduce localization errors, leading to data misalignment during aggregation and significantly degrading cooperative perception performance. Considerable efforts have been made to address these challenges. When2com~\cite{when2com} introduced a handshake mechanism to select the most relevant collaborators for cooperation. V2VNet~\cite{v2vnet} proposed a spatially aware graph neural network (GNN) to aggregate the information received from all the nearby vehicles. Where2comm~\cite{hu2022where2comm} generated a confidence feature map using a classification head and randomly selected the top$k$ features for sharing with other agents. However, they do not consider that when $k$ is too large, redundant background features may be selected, thereby increasing the bandwidth burden and reducing accuracy. Conversely, when $k$ is too small,  regions with low confidence scores but containing target objects may be overlooked. To address localization errors, MRCNet~\cite{mrcnet} proposed Multi-scale Robust Fusion (MRF), which employs cross-semantic, multi-scale enhanced aggregation to fuse features at different scales.

However, the aforementioned methods do not fully exploit the prior knowledge embedded within the feature map. To address this gap, we propose a cooperative perception method based on prior knowledge, termed \textit{Fast2comm}. As illustrated in Fig.~\ref{fig_1:model_overall}, \textit{Fast2comm} achieves effective and efficient feature fusion through three key modules: (1):~Confidence Feature Generation module: To address the problem that the generated confidence feature map does not accurately represent the spatial position of targets, we propose a confidence feature generation method based on prior supervision. This approach ensures that the resulting confidence feature map clearly distinguishes between foreground and background, facilitating feature selection for sharing. (2):~GT Bbox-Based Feature Selection module:~To mitigate issues of redundant or insufficient feature sharing, we propose a GT Bounding Box-based spatial prior feature selection strategy. By selecting features within a predefined BEV bounding box, this method ensures that critical prior information is captured, achieving a balance between accuracy and bandwidth efficiency while enhancing robustness to localization errors. (3):~Feature Fusion module:~To fully integrate the received complementary features, we concatenate the confidence features with those selected based on the GT Bounding Box. Additionally, we decouple the feature fusion strategies during training and testing phases to further optimize bandwidth usage. Extensive experiments on both real-world and simulated datasets demonstrate that our method achieves an effective bandwidth-accuracy trade-off under various localization error conditions.

The main contributions of this work are summarized as follows:
\begin{enumerate}
	\item We propose \textit{Fast2comm}, a communication-efficient and robust multi-vehicle perception framework. The methods introduced in this work effectively address the challenges of communication bandwidth constraints and localization errors.
	\item We develop a confidence feature generation method based on prior supervision and a GT Bounding Box-based feature selection method, leveraging spatial prior knowledge to improve perception performance and enhance robustness against localization errors.
	\item We conduct extensive experiments on both real-world and simulated datasets. The experimental results demonstrate the superior performance of our approach and validate the effectiveness of the proposed methods.
\end{enumerate}

\section{Related works}
\subsection{Multi Agent communication}
Efficient communication plays a critical role in multi-agent systems. Early multi-agent communication methods~\cite{auction,multiagent} relied on predefined protocols and heuristic approaches to regulate interactions between agents. However, these fixed strategies are inadequate for complex and dynamic environments. Consequently, recent research has focused on learning-based approaches to address communication in more challenging scenarios. MAGIC~\cite{magic} employs graph attention mechanisms to determine when and to whom messages should be transmitted. The work in~\cite{communication_itsc} proposes two novel communication protocols based on multi-agent reinforcement learning: the first protocol does not incorporate explicit semantics, serving as a baseline for performance, while the second protocol integrates the concept of advantageous directions, embedding semantic information into communication to enhance interpretability. Due to the absence of explicit supervision, most previous studies concentrate on decision-making tasks and primarily rely on reinforcement learning. In this work, we propose supervising the feature maps used in communication to ensure that the generated confidence map accurately represents the spatial position of targets and clearly distinguishes between foreground and background.
\subsection{Collaborative perception}
Cooperative perception focuses on aggregating complementary perceptual semantics among agents to improve overall system performance. With the availability of new cooperative perception datasets~\cite{opv2v,v2xvit,DAIRV2X,SCOPEdata}, several research efforts have emerged. Method~\cite{DFS} proposed to dynamically reduce the feature data required for sharing among the cooperating entities by filtering the feature data based on the designed priority values. Where2comm\cite{hu2022where2comm} shares the top $k$ confidence features with other agents, but their performance is susceptible to the value of $k$. To address this limitation, we propose a GT Bounding Box-based feature selection method that ensures features containing key prior information are shared among agents. Our method achieves a balance between accuracy and bandwidth efficiency while also enhancing robustness to localization errors.

\section{Fast2comm: Collaborative perception with prior knowledge}
\label{fast2comm}
\begin{figure*}[thpb]
	\centering
	\includegraphics[width=6in]{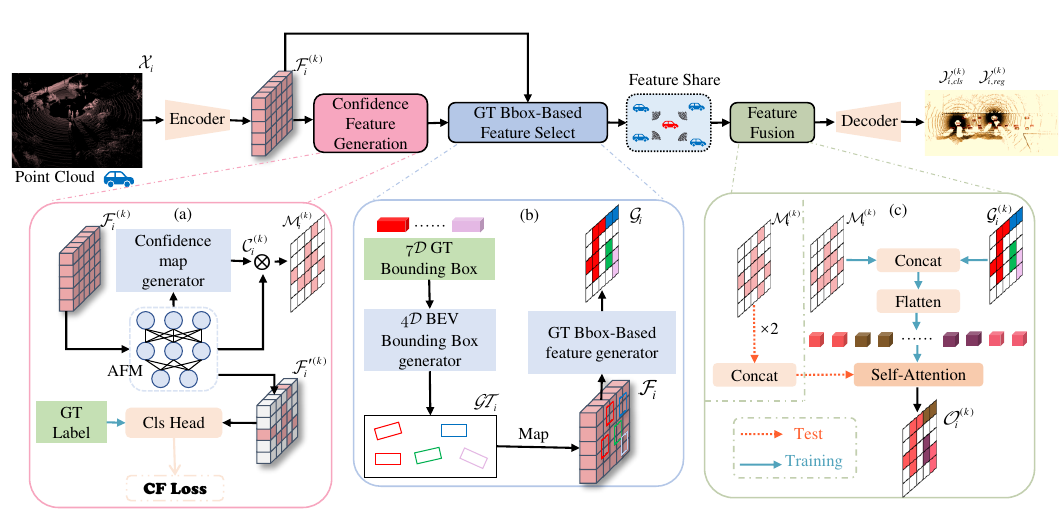}
	\caption{The overall architecture of the proposed \textit{Fast2comm}. The framework consists of six modules: Encoder, Confidence Feature Generation module, GT Bbox-Based Feature Select module, Feature Share, Feature Fusion, and Decoder. The details of each individual component are illustrated in Section~\ref{fast2comm}.}
	\label{fig_1:model_overall}
\end{figure*}

This section introduces \textit{Fast2comm}, a collaborative perception framework based on prior knowledge. Fig.~\ref{fig_1:model_overall} illustrates the overall structure of the proposed framework. \textit{Fast2comm} comprises six sequential components: an Encoder, a Confidence Feature Generation module, a GT Bbox-based Feature Selection module, a Feature Sharing module, a Feature Fusion module, and a Decoder. In the Confidence Feature Generation module, we propose a confidence map generation method based on ground truth supervision. The generated confidence map incorporates spatial prior knowledge, thereby effectively distinguishing between foreground and background regions. In the GT Bbox-Based Feature Selection module, we propose a shared feature selection method based on BEV bounding boxes to address the issues of feature redundancy and insufficient sharing during multi-agent communication, achieving a balance between communication bandwidth efficiency and detection performance. Additionally, it enhances the model's robustness to positional errors.

\subsection{Encoder}
Like most collaborative perception models, \textit{Fast2comm} encodes 3D point clouds into Bird's Eye View (BEV) features to extract local visual representations. Given the local observations $\mathcal{X}_i$ of the $i$-th agent, the extracted feature map is denoted as $\mathcal{F}_i^{(0)} = f_{enc}\left(\mathcal{X}_i\right) \in \mathbb{R}^{C\times H\times W}$, where $f_{enc}$ represents the PointPillar\cite{lang2019pointpillars} encoder shared by all agents. The superscript $(0)$ indicates that the feature is obtained before sharing, and $C$, $H$, and $W$ represent channel, height, and width. The extracted feature maps are then fed into the Confidence Feature Generation module and the GT Bbox-Based Feature Selection module.

\subsection{Confidence Feature Generation}
\label{sec:Confidence Feature Generation}
Previous studies have utilized elaborate mechanisms such as spatial heterogeneity map\cite{hu2022where2comm,zhang2024ermvp} to balance accuracy and required transmission bandwidth. However, these methods generate spatial heterogeneity maps directly from local visual representations, ignoring the prior knowledge of ground truth labels. As a result, the generated confidence map cannot effectively reflect the actual location and confidence score of the object in space. To bridge the gap, we introduce an advanced prior-based spatial confidence feature-generating strategy, see Fig.~\ref{fig_1:model_overall}(a). 

We first employ an Attention Fusion Module (AFM) to aggregate the feature map $\mathcal{F}_{i}^{(k)}$, resulting in the fused feature $\mathcal{F'}_{i}^{(k)}$. The aggregated feature $\mathcal{F'}_{i}^{(k)}$ is then fed into the confidence map generator, where a classification head separately produces the confidence map and the prediction results. In the following sections, we will introduce the design of the Confidence Map Generator and the Attention Fusion Module in detail.

\subsubsection{Attention Fusion Module and Prior Supervision}
\begin{figure}[thpb]
	\centering
	\includegraphics[width=3in]{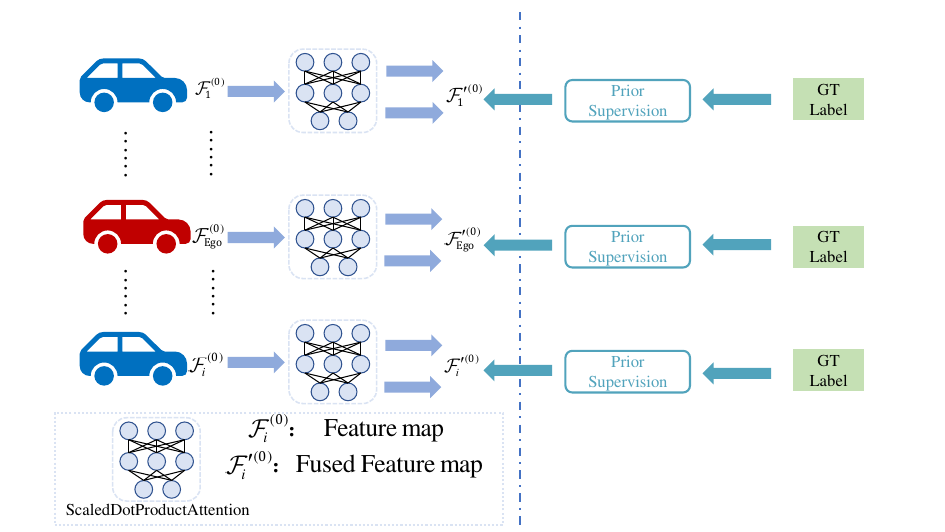}
	\caption{The process of the proposed attention fusion module and prior supervision.}
	\label{fig_2:AFM}
\end{figure}

We utilized ScaledDotProductAttention\cite{vaswani2017attention} to fuse the features $\mathcal{F}_{i}^{(0)}$ of each agent, generating the aggregated feature $\mathcal{F'}_{i}^{(0)}$, as illustrated in Fig.~\ref{fig_2:AFM}. Notably, $\mathcal{F'}_{i}^{(0)}$ is supervised by GT labels. To implement this strategy, $\mathcal{F'}_{i}^{(0)}$ is passed through a classification head to produce a feature map of size $2 \times H \times W$. The prior loss is then computed by comparing the generated feature map with the GT labels. By incorporating ground truth supervision, $\mathcal{F'}_{i}^{(0)}$ embeds rich prior information, thereby ensuring higher accuracy when generating the spatial confidence map.
\subsubsection{Confidence Map Generator}
Intuitively, in object detection tasks, foreground areas containing objects are more important than background areas. During collaborative perception, foreground areas with objects can help restore the miss-detected objects due to occlusion and limited view, while background regions can be omitted to save communication bandwidth.

Following Where2comm~\cite{hu2022where2comm}, the spatial confidence map is represented by the detection confidence map, where areas with higher perceptual criticality correspond to regions containing objects with high detection confidence scores. But different from where2comm, \textit{Faste2comm} uses the aggregated feature $\mathcal{F'}_{i}^{(k)}$ which integrates prior foreground knowledge, as the input to the confidence map generator. Given the feature map at the $K$th communication round, $\mathcal{F'}_{i}^{(k)}$, the corresponding spatial confidence map is defined as:
\begin{equation}
	\mathcal{C}_{i}^{(k)} = \Psi_{t} \left(\Theta_{generator}\left(\mathcal{F'}_{i}^{(k)}\right) \right) \in \left\{0,1\right\}^{H\times W}
\end{equation}
where $\Theta_{generator}$ stands for detection decoder, $\Psi_{t}$ indicates thresholding the confidence map with threshold $t$. The confidence map $\mathcal{C}$ represents whether each spatial location is selected, where $1$ denotes a selected location and $0$ otherwise. Moreover, because $\mathcal{F'}$  contains abundant prior clues, the generated confidence map can more accurately localize the position of the target in space.

After obtaining the confidence map $\mathcal{C}$, we perform an element-wise multiplication between the feature map $\mathcal{F}$ and the confidence map $\mathcal{C}$ to produce a spatially sparse yet perceptually critical feature map:
\begin{equation}
	\mathcal{M}_{i}^{(k)} = \mathcal{C}_{i}^{(k)} \otimes \mathcal{F}_{i}^{(k)}
\end{equation}
where $\otimes$ reprensents the element-wise multiplication.

\subsection{GT Bounding Box-Based Feature Select}
\label{sec:GT Bounding Box-Based Feature Select}
During training, when communicating with other agents, existing methods\cite{hu2022where2comm,zhang2024ermvp} randomly selects the top-$k$ values from the feature map $\mathcal{M}_{i}$ for sharing. However, when $k$ is too large, it results in the transmission of a significant amount of irrelevant information, introducing noise and increasing bandwidth consumption. Conversely, when $k$ is too small, important features may be overlooked. Moreover, randomly selecting maximum values does not consider the spatial context or local structure within different regions of the feature map, thereby limiting the comprehensiveness of the environmental perception provided to other agents.
To address these limitations, we propose a GT Bounding box-based spatial prior feature selection method, which ensures that regions containing targets are selected for sharing. This approach enables the network to focus on key target features while reducing interference from background information. The process is illustrated in Fig.\ref{fig_1:model_overall}(b).

We first obtain the 7D bounding box of each object projected into the ego coordinate system, denoted as $(x, y, z, l, w, h, \theta)$, where $(x, y, z)$ represent the center coordinates of the object in 3D space, $(l, w, h)$ correspond to its length, width, and height, and $\theta$ denotes the heading angle. Based on the 7D bounding box, we further derive the corresponding 4D bounding box in the Bird's Eye View (BEV) space using a BEV Bounding Box Generator.
\subsubsection{4D BEV Boudning Box Generator}
\begin{figure}[thpb]
	\centering
	\includegraphics[width=3in]{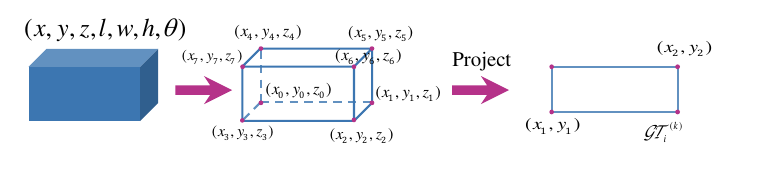}
	\caption{The process of the proposed 4D bounding box generating.}
	\label{fig_3:4Dboundingboxgenerator}
\end{figure}

The process is illustrated in Fig.~\ref{fig_3:4Dboundingboxgenerator}. Specifically, we first convert the 7D bounding box in the world coordinate system into eight corner points of the corresponding cuboid, resulting in a set of coordinates with dimensions $(8, 3)$. These 3D coordinates are then projected onto the 2D BEV plane to generate the 4D bounding box $\mathcal{GT}_i$, where $(x_1', y_1')$ denotes the bottom-left corner and $(x_2', y_2')$ denotes the top-right corner. It is important to note that the 4D bounding box coordinates are defined in the ego-centered coordinate system.

\subsubsection{GT Bbox-Based Feature Generator}
\begin{figure}[thpb]
	\centering
	\includegraphics[width=3in]{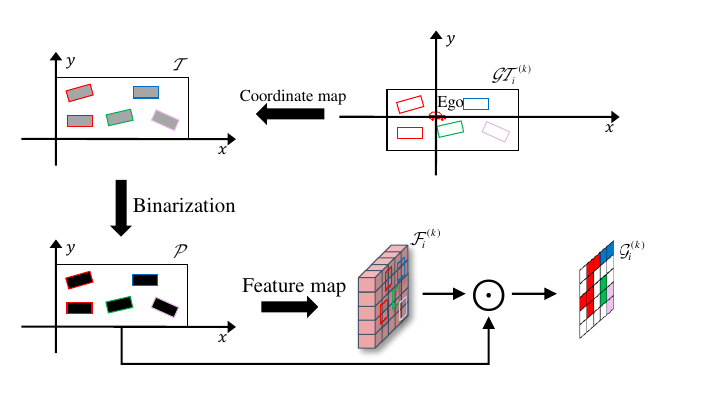}
	\caption{The process of the proposed GT Bbox-Based feature generator.}
	\label{fig_4:GT Bbox-Based feature generator}
\end{figure}

We map the 4D bounding box $\mathcal{GT}_{i}$ onto the feature map $\mathcal{F}_{i}$, selecting features enriched with prior information for sharing, as shown in Fig.~\ref{fig_4:GT Bbox-Based feature generator}.
Specifically, we first convert the GT coordinates from the ego-centered coordinate system to the  feature map-based coordinates:
\begin{equation}
	\begin{aligned}
		&x_1=\left(x_1'+r_x\right)/\frac{r_x}{H}, y_1=\left(y_1'+r_y\right)/\frac{r_y}{W} \\
		&x_2=\left(x_2'+r_x\right)/\frac{r_x}{H}, y_2=\left(y_2'+r_y\right)/\frac{r_y}{W} 
	\end{aligned}
\end{equation}
where $(x_1', y_1')$ and $(x_2', y_2')$ represent the coordinates in the ego-centered coordinate system, $(x_1, y_1)$ and $(x_2, y_2)$ represent the coordinates in the feature map-based coordinates, $r_x$ and $r_y$ denote the detection range of the LiDAR in the x and y directions, and $H$ and $W$ represent the height and width of the feature map, respectively.

Then, given a tensor $\mathcal{T}_i$ initialized with zeros and having the same spatial dimensions as $\mathcal{F}_i$, we set the values within the coordinate range defined by $x_2 - x_1$ and $y_2 - y_1$ in $T$ to 1, resulting in the prior knowledge binary map $\mathcal{P}_i$:
\begin{equation}
	\begin{aligned}
		\mathcal{T}:
		&\begin{pmatrix}
			0_{a_{11}} & \cdots & 0_{a_{1W}} \\
			\vdots & \ddots & \vdots \\
			0_{a_{H1}} & \cdots & 0_{a_{HW}}
		\end{pmatrix} \rightarrow \\
		\mathcal{P}:
		&\begin{pmatrix}
			0_{a_{11}} & \cdots & 0_{a_{1W}} \\
			\vdots & \begin{pmatrix}
				1_{(a_{x_1},a_{y_2})} &\cdots & 1_{(a_{x_2},a_{y_2})} \\
				\vdots & \ddots & \vdots \\
				1_{(a_{x_1},a_{y_1})} & \cdots & 1_{(a_{x_2},a_{y_1})}
			\end{pmatrix} & \vdots \\
			0_{a_{H1}} & \cdots & 0_{a_{HW}}
		\end{pmatrix}
	\end{aligned}
\end{equation}
where $a_{HW}$ represents the feature map coordinate index. The regions in $\mathcal{P}_i$ where the values are equal to 1 indicate the locations of the objects.

After obtaining $\mathcal{P}_i$, the proposed GT Bbox-Based Feature Generation method maps $\mathcal{P}_i$ onto the feature map $\mathcal{F}_i$ by performing element-wise multiplication $\mathcal{P}_i \odot \mathcal{F}_i$, resulting in the generated feature $\mathcal{G}_i$. Since $\mathcal{G}_i$ incorporates rich prior features from the target regions, it ensures that the key information from agent $i$ is shared with the ego agent, effectively removing irrelevant background noise and reducing the interference caused by background features. This process enhances object detection accuracy and improves robustness against positional GPS errors.

\subsection{Feature Share}
In the feature sharing stage, we package the shared information $\mathcal{M}_{i}^{(k)}$ and $\mathcal{G}_{i}$ into a unified message tensor $\mathcal{Z}_{i}^{(k)}$. Overall, the message sent from the $i$th agent to the $j$th agent at the $K$th communication round is represented as: $\mathcal{Z}_{i\rightarrow j}^{(k)}=\left(\mathcal{M}_{i\rightarrow j}^{(k)}, \mathcal{G}_{i\rightarrow j}\right)$. Note that: (1)The feature map $\mathcal{Z}_{i}^{(k)}$ provides supporting information specifically tailored to the needs of agent $i$ during the current communication round, enabling mutually beneficial collaboration; (2) Since $\mathcal{Z}_{i\rightarrow j}^{(k)}$ is spatially sparse, we transmit only the non-zero features along with their corresponding indices, thereby significantly reducing communication costs; (3) The sparsity of $\mathcal{Z}_{i\rightarrow j}^{(k)}$ is controlled by a binary selection matrix, which dynamically allocates the communication budget based on the perception perceptual criticality of each spatial region, thereby adapting to different communication conditions.
\subsection{Feature Fusion}
\label{sec:Feature Fusion}
After receiving the spatially sparse yet perceptually critical features $\mathcal{M}_{i\rightarrow j}^{(k)}$ and $\mathcal{G}_{i\rightarrow j}$ from other agents, the ego agent concatenates $\mathcal{M}_{i\rightarrow j}^{(k)}$ and $\mathcal{G}_{i\rightarrow j}$ for feature fusion. We choose concatenation rather than direct addition for the following reason: $\mathcal{M}_{i\rightarrow j}^{(k)}$ contains the features with the highest confidence scores extracted from the feature map $\mathcal{F}_{i}^{(k)}$, while $\mathcal{G}_{i\rightarrow j}$ contains the features from the target region. Directly adding $\mathcal{M}_{i\rightarrow j}^{(k)}$ and $\mathcal{G}_{i\rightarrow j}$ would result in significantly larger values in the target areas, which would weaken the contribution of regions with fewer features but still containing valuable target information. Subsequently, the concatenated tensor $\mathcal{Z}_i^{(k)}$ is flattened and fed into a Self-Attention module~\cite{vaswani2017attention} to fuse corresponding features received from other agents. By integrating both confidence maps and prior knowledge maps, the ego agent's feature representation is effectively enhanced. The process is illustrated in Fig.~\ref{fig_1:model_overall}(c), and can be expressed as the following formula:

\begin{equation}
	\mathcal{O}_{i}^{(k)}=Self\_Attn\left(Flatten\left(\mathcal{M}_{i\rightarrow j}^{(k)}\cup \mathcal{G}_{i\rightarrow j}\right)\right)
\end{equation}
where $\mathcal{O}_{i}^{(k)}$ is the fused output, $\cup$ repensents the concatenation operation, $k$ repensents the $k$th round of communication.

It is worth noting that \textit{Fast2comm} shares both prior features $\mathcal{G}$ and confidence features $\mathcal{M}$ during training, but only shares confidence features $\mathcal{M}$ during testing. This design improves detection accuracy while reducing communication bandwidth during testing.
\subsection{Decoder}
The decoder decodes feature $\mathcal{O}_{i}^{(k)}$ into the predicted outputs: $\mathcal{Y}_{i,cls}^{(k)},\mathcal{Y}_{i,reg}^{(k)}=\Phi_{decoder}\left(\mathcal{O}_{i}^{(k)}\right)$. The classification output reveals the confidence values for each predefined box as either a target or background, which is $\mathcal{Y}_{i,cls}^{(k)}\in \mathbb{R}^{2\times H\times W}$. The regression output is $\mathcal{Y}_{i,reg}^{(k)}\in \mathbb{R}^{7\times H\times W}$, with $\left(x,y,z,l,w,h,\theta\right)$representing the position, size, and yaw angle of the bounding box.
\subsection{Training Details and Loss Functions}
To ensure that the generated confidence map incorporates spatial prior knowledge, we introduce an additional loss function $L_{pk}$ to supervise its generation. Consequently, \textit{Fast2comm} is supervised by three loss functions, namely $L_{pk}$, $L_{cls}$, and $L_{reg}$. $L_{pk}$ is the prior knowledge loss, $L_{cls}$ is the classification loss, and the $L_{reg}$ is hte regression loss. Following existing work~\cite{lang2019pointpillars}, we adopt the smooth L1 loss for  bounding boxes regression and the focal loss~\cite{focalloss} for both classification and prior knowledge supervision. We use the parameters $\alpha$, $\beta$, and $\gamma$ to balance the importance of each loss. Therefore, the total loss of the model is formulated as:
\begin{equation}
	\mathcal{L}_{total} = \alpha \cdot \mathcal{L}_{cls} + \beta \cdot \mathcal{L}_{reg} + \gamma \cdot \mathcal{L}_{pk}
\end{equation}
where $\mathcal{L}_{cls}$, $\mathcal{L}_{reg}$, and $\mathcal{L}_{pk}$ represent the individual loss functions, and $\alpha$, $\beta$, and $\gamma$ are the corresponding weights.

\section{Experimental Results}
\subsection{Datasets and Evaluation Metrics}
\subsubsection{Datasets}
We validate the effectiveness of the proposed model on three public datasets: OPV2V~\cite{opv2v}, V2XSet~\cite{v2xvit}, and DAIR-V2X~\cite{DAIRV2X}.
\textbf{OPV2V} is a large-scale vehicle-to-vehicle cooperative perception dataset comprising 73 diverse scenarios, each involving 2 to 7 cooperating vehicles. Each vehicle is equipped with a LiDAR sensor and four cameras. The dataset includes 11,464 frames of point clouds and RGB images, split into 6,374 training frames, 1,980 validation frames, and 2,170 test frames.
\textbf{V2XSet} is a publicly available simulated dataset for vehicle-to-everything (V2X) cooperative perception. It provides 73 representative scenarios and 11,447 annotated point cloud frames, generated using CARLA~\cite{carla}. The training, validation, and test sets consist of 6,694, 1,920, and 2,833 frames, respectively.
\textbf{DAIR-V2X} is a large-scale real-world dataset for cooperative 3D object detection, containing 71,254 samples. It is split into training, validation, and test sets according to a 5:2:3 ratio. Each sample includes LiDAR point clouds collected from both vehicles and roadside infrastructure sensors.
\subsubsection{Evaluation metrics}
We evaluate the 3D object detection performance using Average Precision (AP) at Intersection over Union (IoU) thresholds of 0.5 and 0.7. The communication cost is measured in bytes, and the message size is reported using a base-2 logarithmic scale to reflect transmission efficiency.
\subsection{Implementation Details}
We implemented the proposed \textit{Fast2comm} model and its baselines using PyTorch~\cite{pytorch}, and trained them on two NVIDIA GeForce RTX 3090 GPUs with the Adam optimizer~\cite{adam}. The initial learning rate was set to $2 \times 10^{-4}$ and scheduled using a cosine annealing strategy. All models were trained for 60 epochs with a batch size of 4.
We applied standard point cloud data augmentation techniques, including random scaling, rotation, and flipping, to all experiments. All detection models are based on the PointPillars~\cite{lang2019pointpillars} backbone, which extracts 2D features from point clouds. The width and length of each voxel were set to 0.4 meters.
To simulate localization errors, we added Gaussian noise with a standard deviation of $\sigma_e$ to the positional data during both training and evaluation.

\subsection{Quantitative Evaluation}
\subsubsection{Benchmark Comparison} 
Table~\ref{tab:table1} summarizes the 3D object detection results on the three datasets. Compared to the baseline model Where2comm~\cite{hu2022where2comm}, \textit{Fast2comm} achieves improvements of 1.0\%/1.2\% on the OPV2V dataset, 2.7\%/2.9\% on the V2XSet dataset, and 1.5\%/0.9\% on the DAIR-V2X dataset.
Our method attains results comparable to state-of-the-art models Scope~\cite{scope} and MRCNet~\cite{mrcnet} on the OPV2V and V2XSet datasets. Notably, on DAIR-V2X at AP@0.7, \textit{Fast2comm} outperforms Scope~\cite{scope} by 1.9\%. In addition, on OPV2V AP@0.7 and V2XSet AP@0.7, \textit{Fast2comm} surpasses MRCNet~\cite{mrcnet} by significant margins of 3.5\% and 5.3\%, respectively.
These results demonstrate the effectiveness and competitiveness of the proposed method across various cooperative perception benchmarks.
\begin{table}[!t]
	\caption{Performance comparison on the OPV2V, V2XSet, and DAIR-V2X datasets. '?' indicates that the corresponding results were not reported in their paper.\label{tab:table1}}
	\centering
	\begin{tabular}{c|c|c|c}
		\toprule[1pt]
		Model & \begin{tabular}[c]{@{}c@{}}OPV2V \\ AP@0.5/0.7\end{tabular} & \begin{tabular}[c]{@{}c@{}}V2XSet \\ AP@0.5/0.7\end{tabular} & \begin{tabular}[c]{@{}c@{}}DAIR-V2X \\ AP@0.5/0.7\end{tabular}\\
		\midrule
		No Fusion & 68.71/48.66 & 60.60/40.20 & 50.03/43.57 \\
		Late Fusion &  82.24/65.78 &66.79/50.95 & 53.12/37.88 \\
		Early Fusion & 68.71/48.66 & 60.60/40.20 & 50.03/43.57 \\
		When2comm~\cite{when2com} & 77.85/62.40 & 70.16/53.72 & 51.12/36.17 \\
		V2VNet~\cite{v2vnet} & 82.79/70.31 & 81.80/61.35 & 56.01/42.25 \\
		AttFuse~\cite{opv2v} & 83.21/70.09 & 76.27/57.93 & 53.79/42.61 \\
		V2X-Vit\cite{v2xvit} & 86.72/74.94 & 85.13/68.67 & 54.26/43.35 \\
		DiscoNet\cite{disconet} & 87.38/73.19 & 82.18/63.73 & 54.29/44.88 \\
		CoBEVT~\cite{cobevt} & 87.40/74.35 & 83.01/62.67 & 54.82/43.95 \\
		Scope~\cite{scope} & 89.71/80.62 & 87.52/75.05 & 65.18/49.89 \\ 
		How2Comm~\cite{how2comm} & 85.42/72.24 & 84.05/67.01 & 62.36/47.18 \\
		MRCNet~\cite{mrcnet} & 89.77/76.12 & 85.00/66.31 & -/- \\
		Where2comm~\cite{hu2022where2comm} & 87.80/78.44 & 82.04/68.73 & 63.13/50.84 \\
		\textbf{Fast2comm(ours)} & \textbf{88.86/79.62} & \textbf{84.71/71.61} & \textbf{64.81/51.74} \\
		\bottomrule[1pt]
	\end{tabular}
\end{table}
\subsubsection{Comparison of Communication Volume} 
\begin{figure*}[thpb]
	\centering
	\subfloat[]{\includegraphics[width=1.8in]{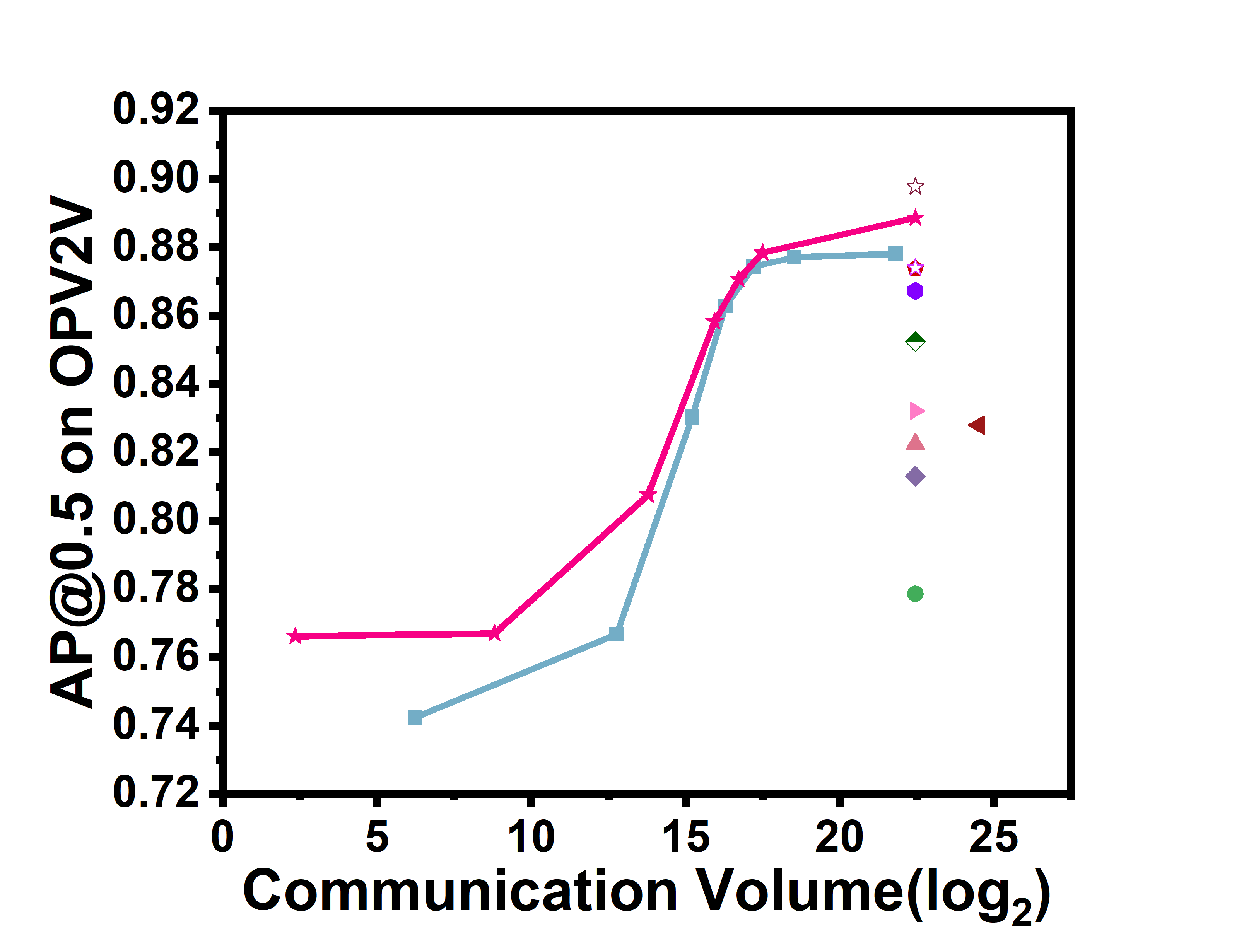}%
		\label{volume_a}}
	\hspace{-6mm}
	\subfloat[]{\includegraphics[width=1.8in]{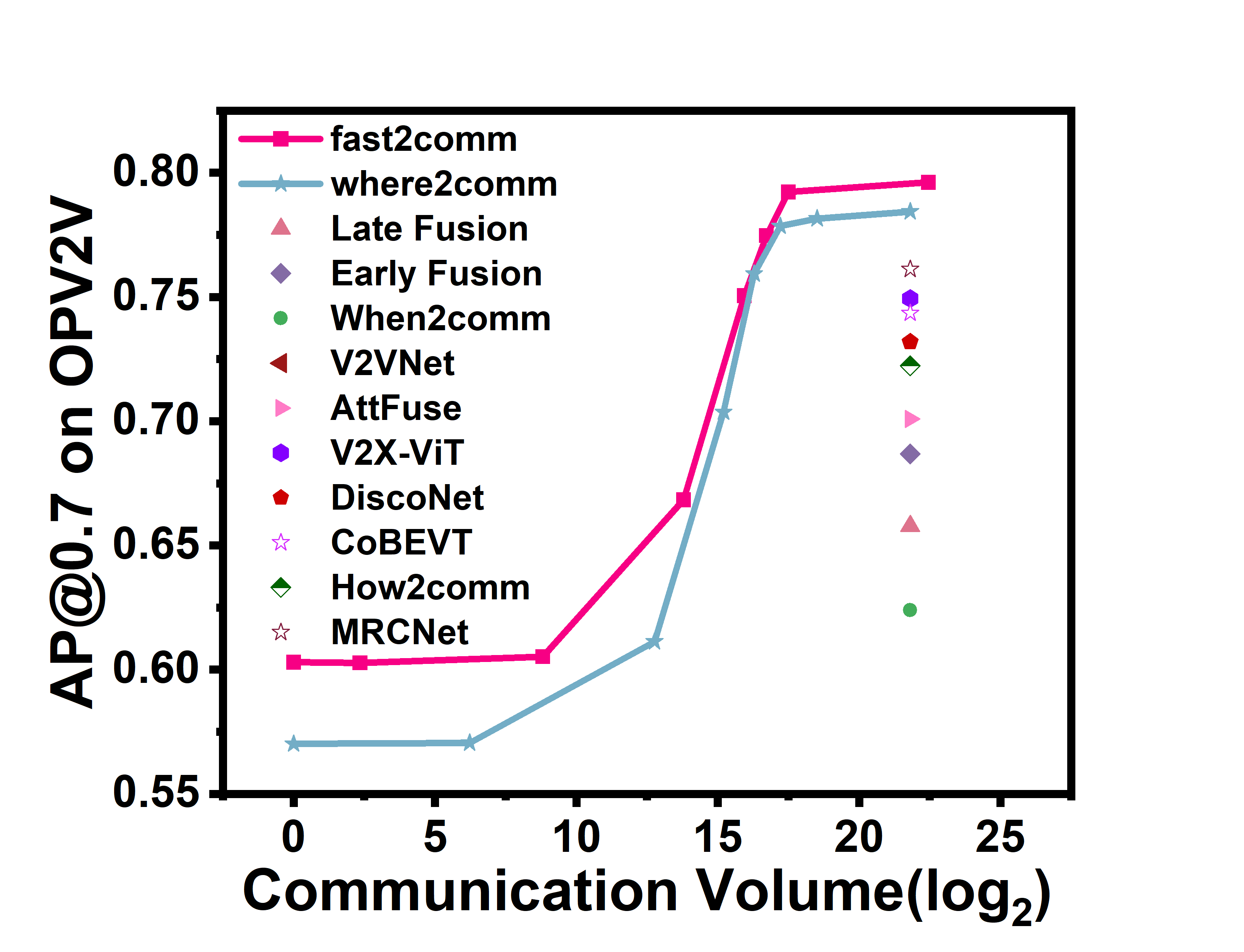}%
		\label{volume_b}}
	\hspace{-6mm}
	\subfloat[]{\includegraphics[width=1.8in]{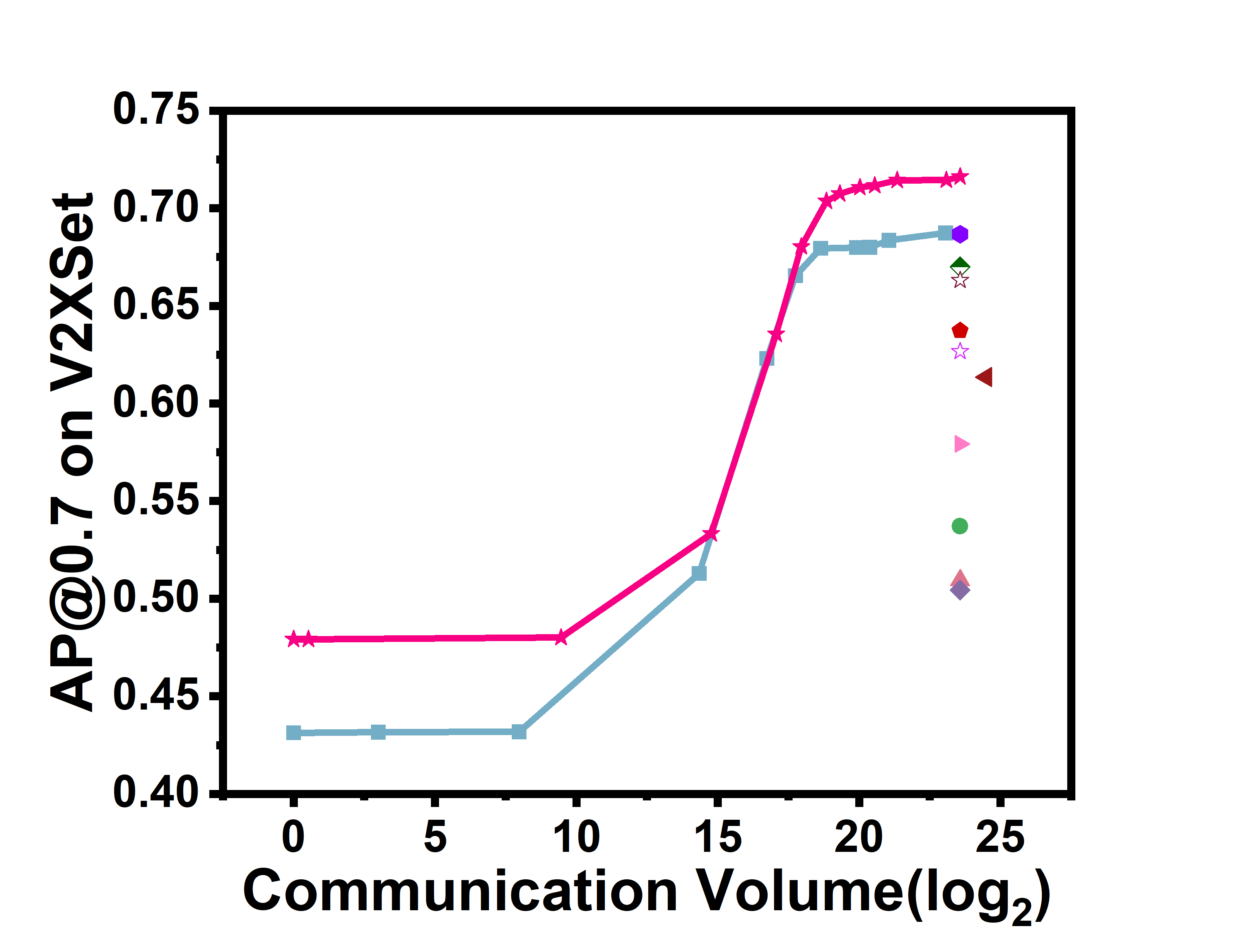}%
		\label{volume_c}}
	\hspace{-6mm}
	\subfloat[]{\includegraphics[width=1.8in]{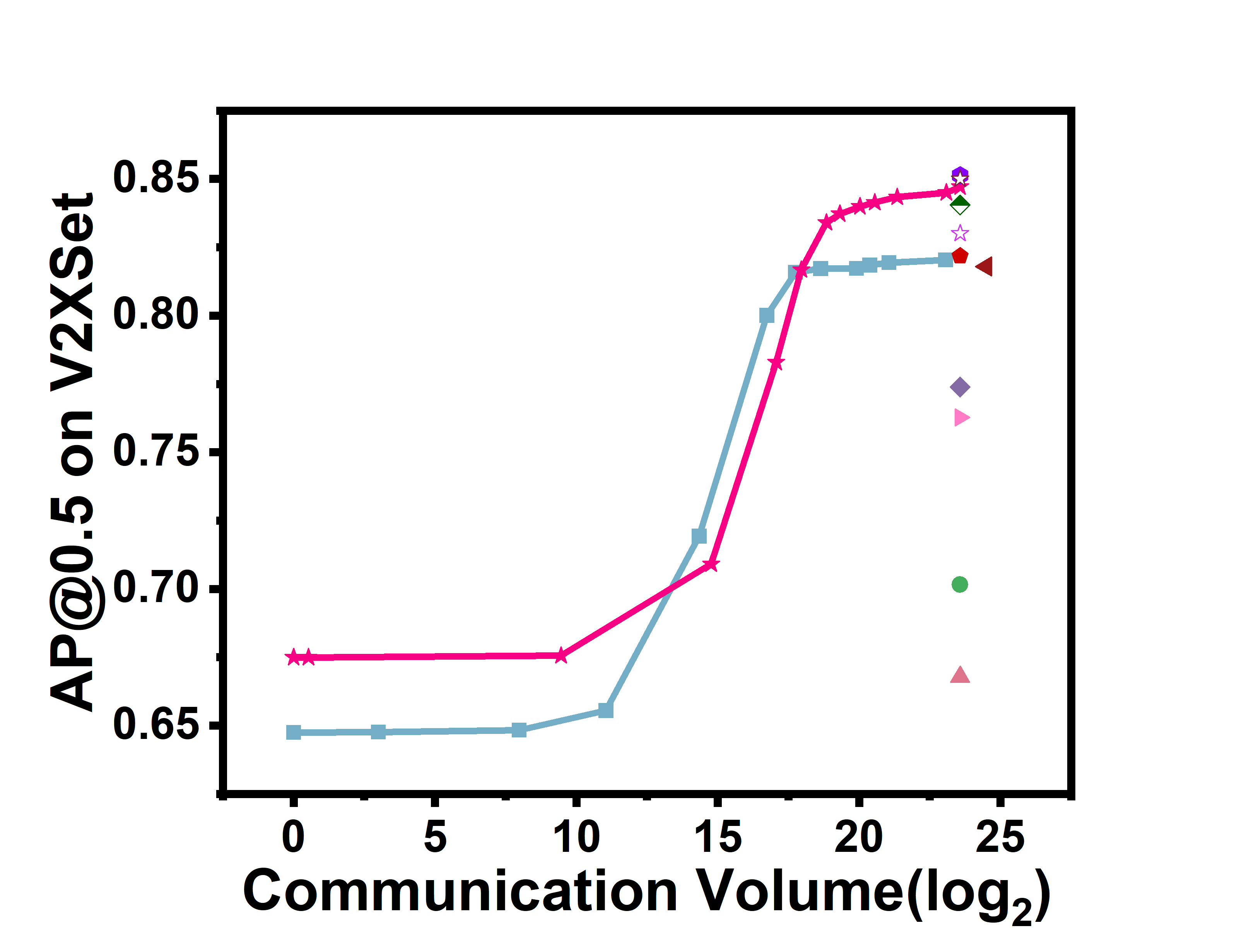}%
		\label{volume_d}}
	\caption{Collaborative perception performance comparison with varying communication volume.}
	\label{fig:communication volume}
\end{figure*}

Figure~\ref{fig:communication volume} illustrates the collaborative perception performance under varying communication volumes. It can be observed that the proposed \textit{Fast2comm}:
(1)~consistently surpasses the baseline model Where2comm~\cite{hu2022where2comm}, achieving a superior perception-communication trade-off across all communication bandwidth settings;
(2)~achieves significant improvements over previous SOTA models~\cite{how2comm,mrcnet,v2xvit,cobevt} on all datasets while requiring less communication volume.
\subsubsection{Robutness to Localization Error} 
\begin{figure*}[thpb]
	\centering
	\subfloat[]{\includegraphics[width=1.8in]{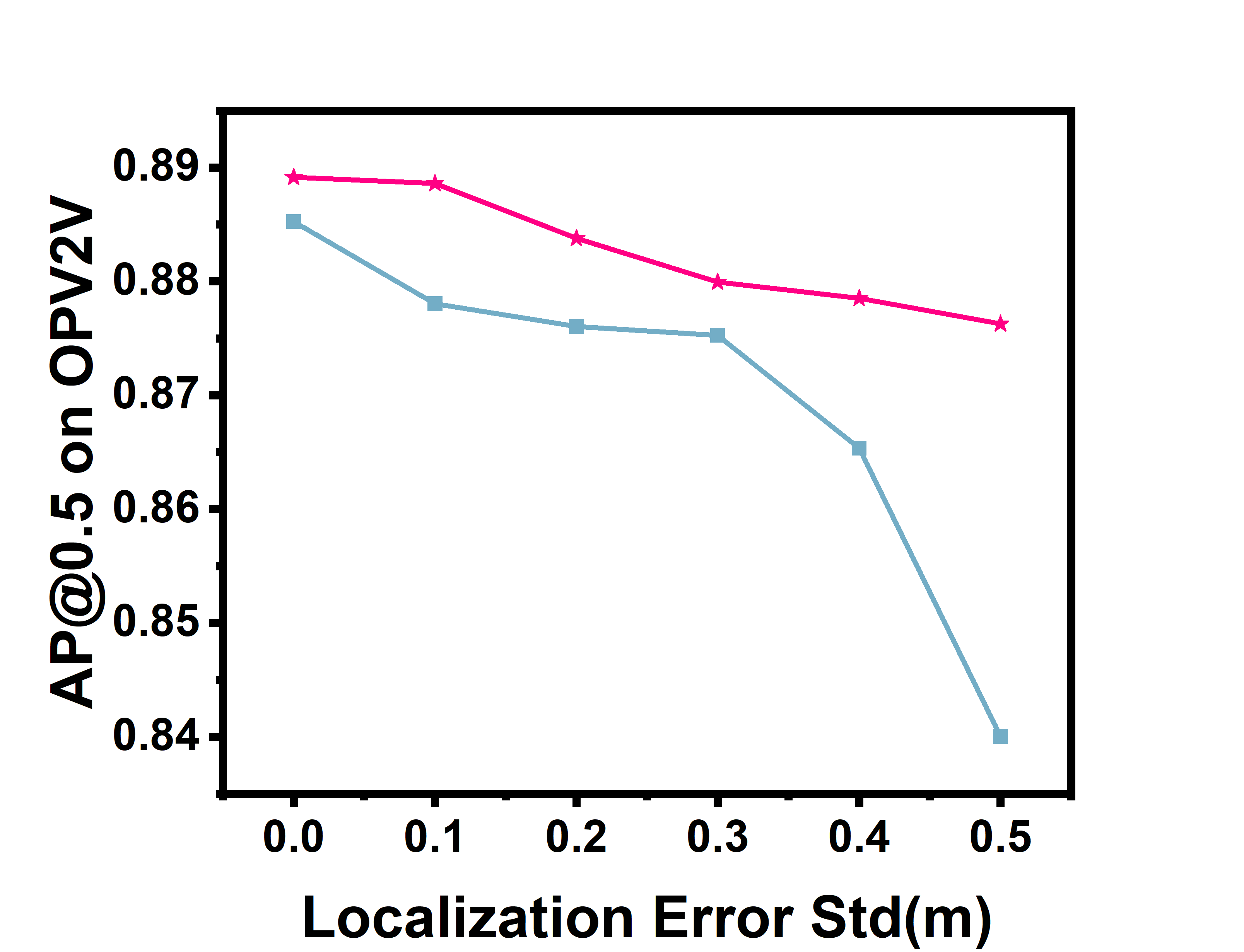}%
		\label{error_a}}
	\hspace{-6mm}
	\subfloat[]{\includegraphics[width=1.8in]{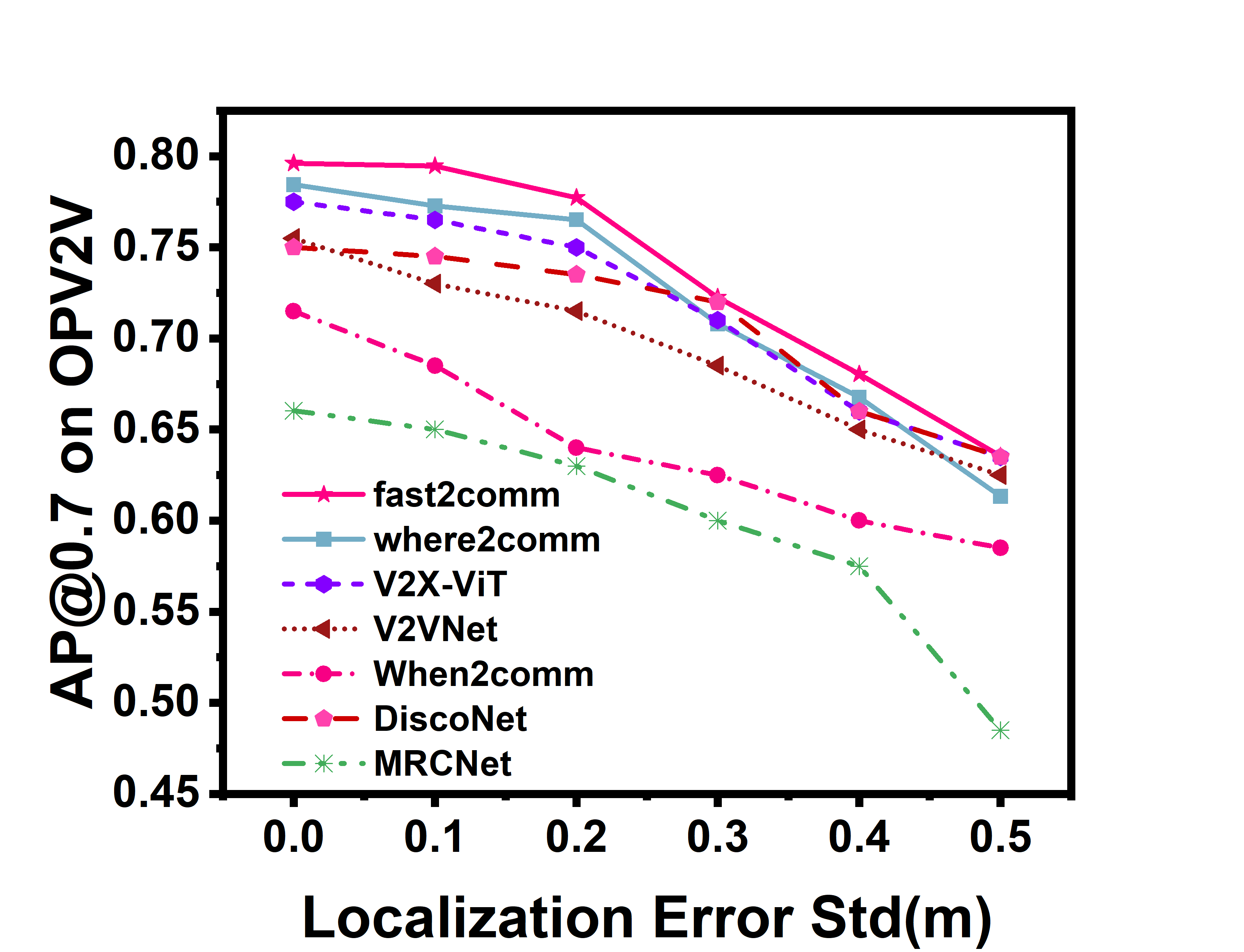}%
		\label{error_b}}
	\hspace{-6mm}
	\subfloat[]{\includegraphics[width=1.8in]{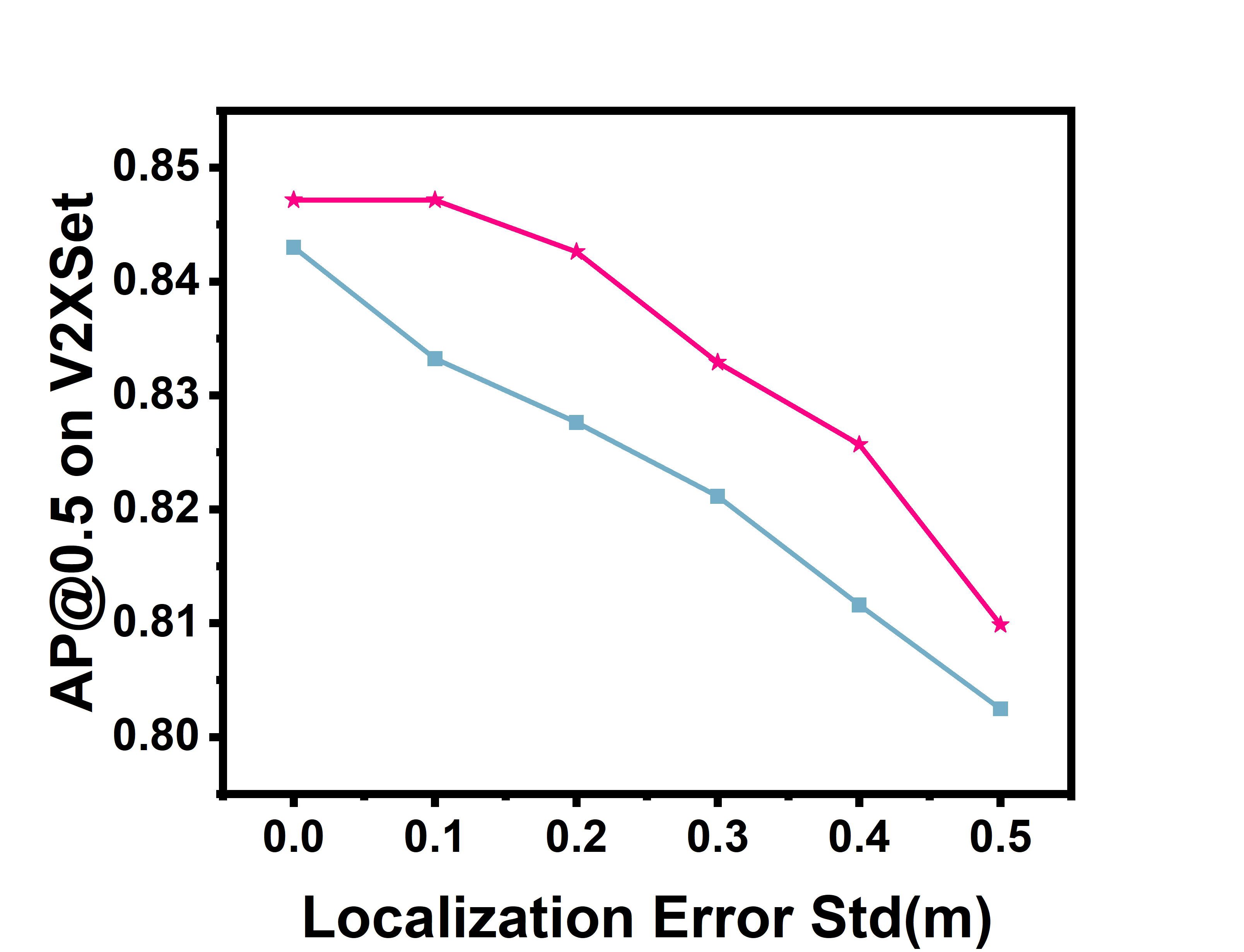}%
		\label{error_c}}
	\hspace{-6mm}
	\subfloat[]{\includegraphics[width=1.8in]{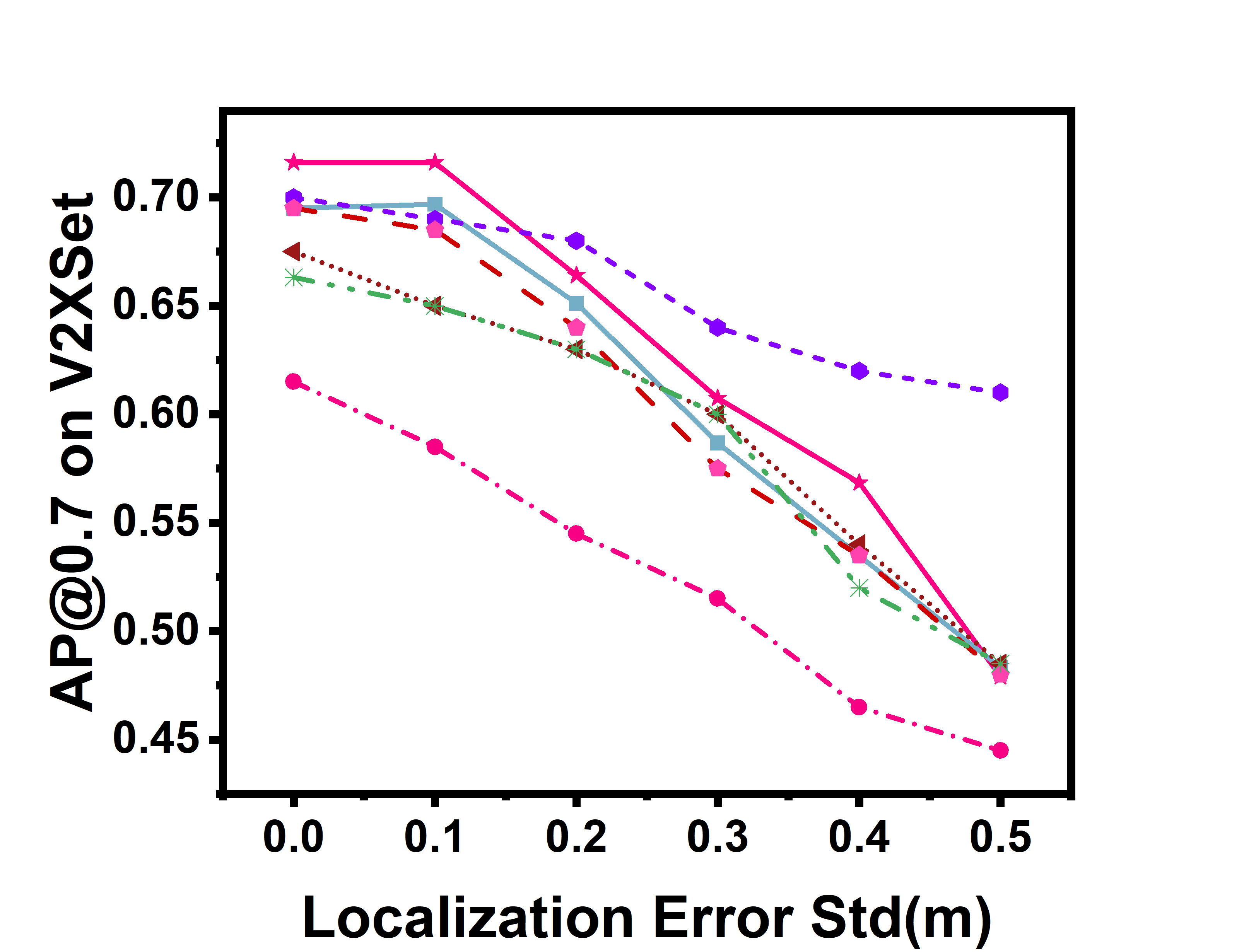}%
		\label{error_d}}
	\caption{Robustness to localization error. \textit{Fast2comm} outperforms baseline model and previous models.} 
	\label{fig:localization error}
\end{figure*}

Figure~\ref{fig:localization error} illustrates the perception performance under different localization errors. The localization noise is sampled from a Gaussian distribution with a standard deviation $\sigma_e \in \left[0, 0.5\right]$. It is noteworthy that our method consistently outperforms other SOTA approaches~\cite{disconet,v2xvit,mrcnet} across all noise levels. Specifically, on the OPV2V dataset at AP@0.5 with a localization error of 0.5 meters, \textit{Fast2comm} achieves a 3.6\% improvement over Where2Comm~\cite{hu2022where2comm}. This robustness can be attributed to the proposed Confidence Feature Generation module, which ensures the accuracy of the generated confidence map, and the GT Bbox-Based Feature Selection module, which selects key prior information through bounding box-based selection. Together, these components facilitate effective interaction among agents and mitigate misalignment in the aggregated feature maps.

\subsection{Qualitative evaluation }
\subsubsection{Visualization of GT Bounding Box-Based Feature Select}
\begin{figure*}[thpb]
	\centering
	\subfloat[]{\includegraphics[width=3in]{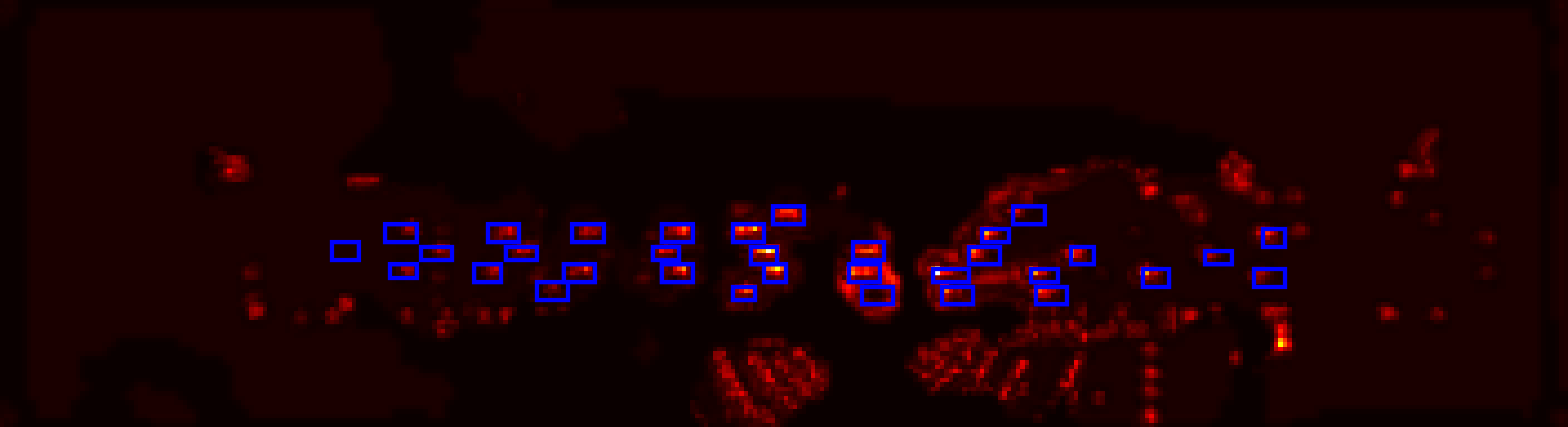}%
		\label{heatmap_a}}
	\hfil
	\subfloat[]{\includegraphics[width=3in]{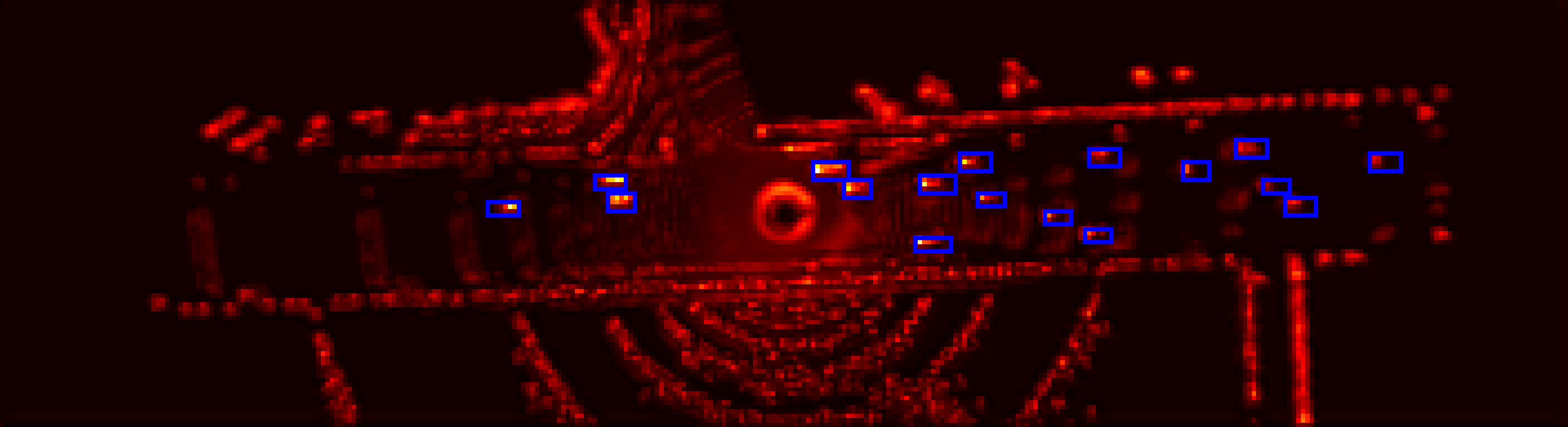}%
		\label{hearmap_b}}
	\caption{Visualization of selected prior features. Brighter regions indicate the locations of the targets, while the blue boxes represent the GT bounding boxes. Fig.~(a) and Fig.~(b) show the visualization results under different scenarios.} 
	\label{fig:hotmap}
\end{figure*}

To verify the effectiveness of the proposed GT Bbox-Based Feature Selection module in selecting key prior features for sharing among agents, Fig.~\ref{fig:hotmap} visualizes the shared feature heatmaps.
It can be clearly observed that our method effectively selects the brightest regions in the feature heatmaps, corresponding to the target areas. These regions contain rich spatial prior knowledge while excluding redundant background information, thereby achieving a favorable balance between perception accuracy and communication bandwidth.
\subsubsection{Visualization of detection results}
\begin{figure*}[thpb]
	\centering
	\subfloat[]{\includegraphics[width=1.8in]{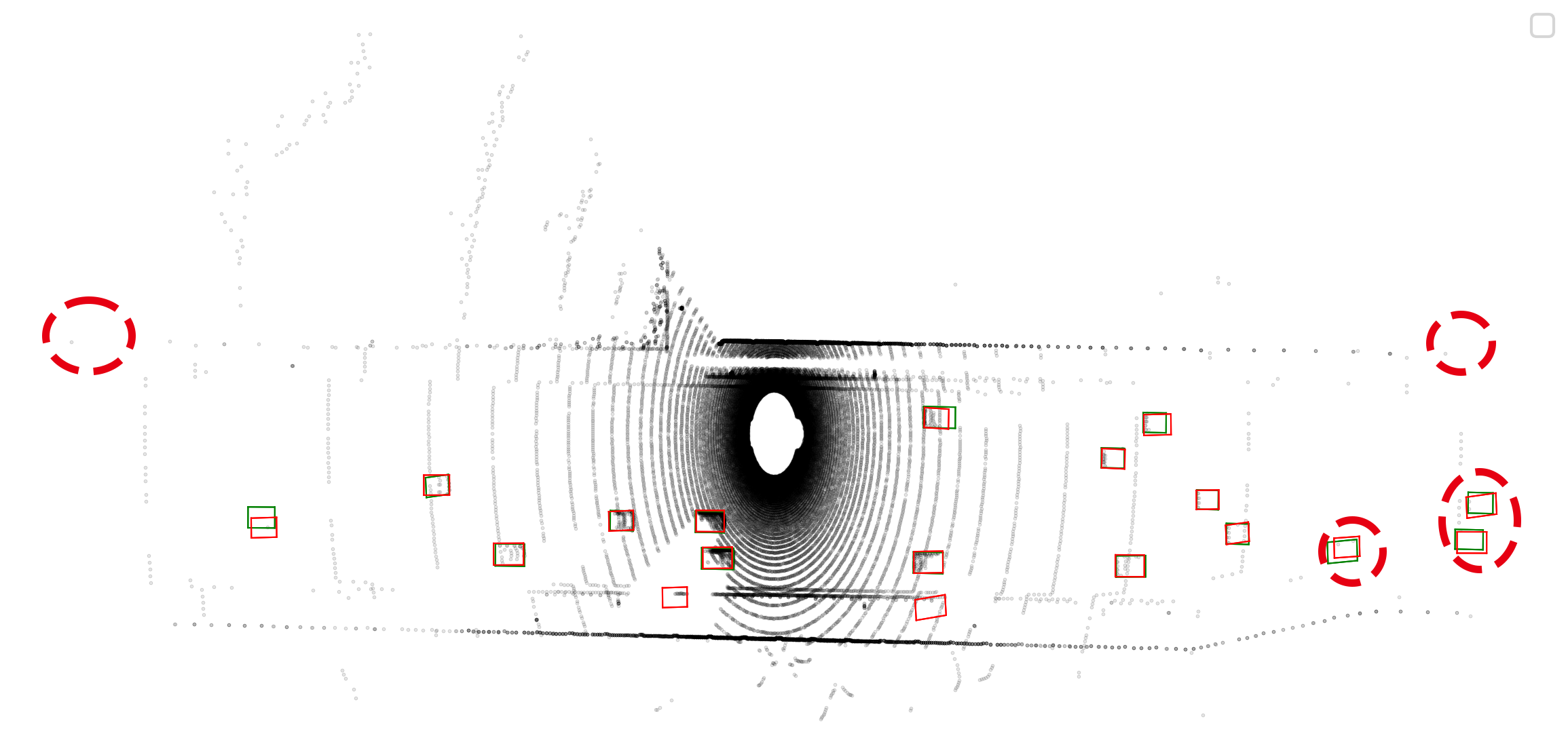} \includegraphics[width=1.8in]{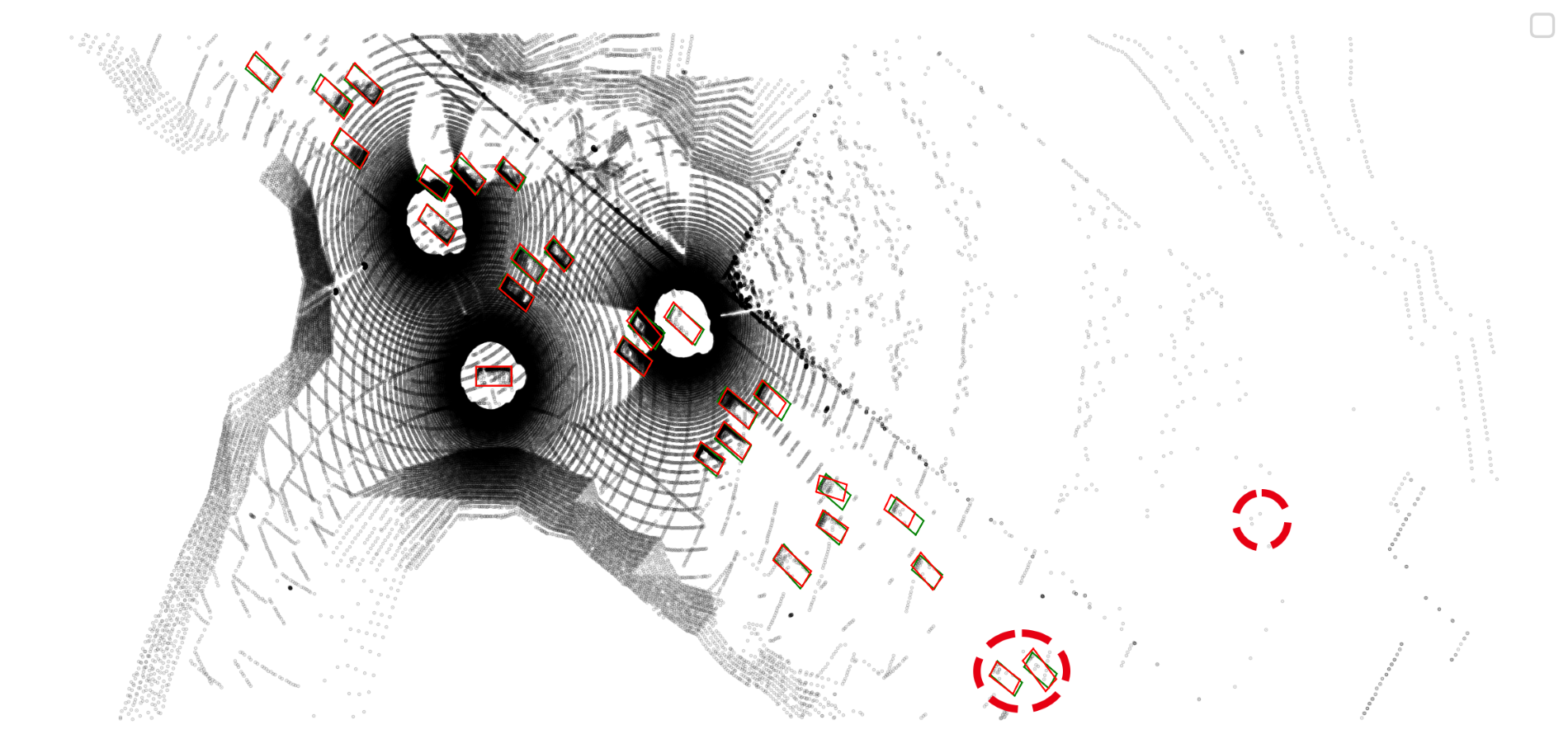}  \includegraphics[width=1.8in]{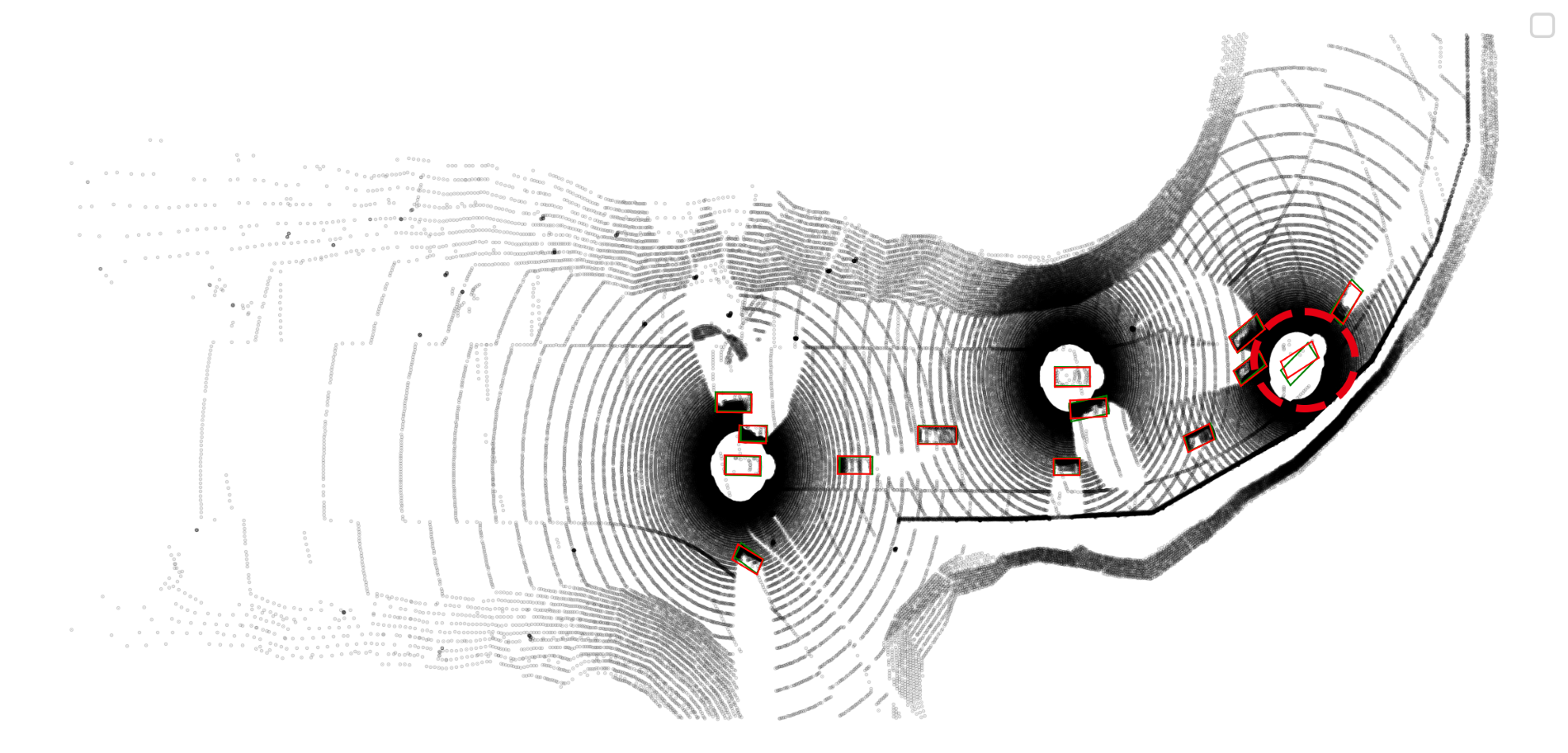} \includegraphics[width=1.8in]{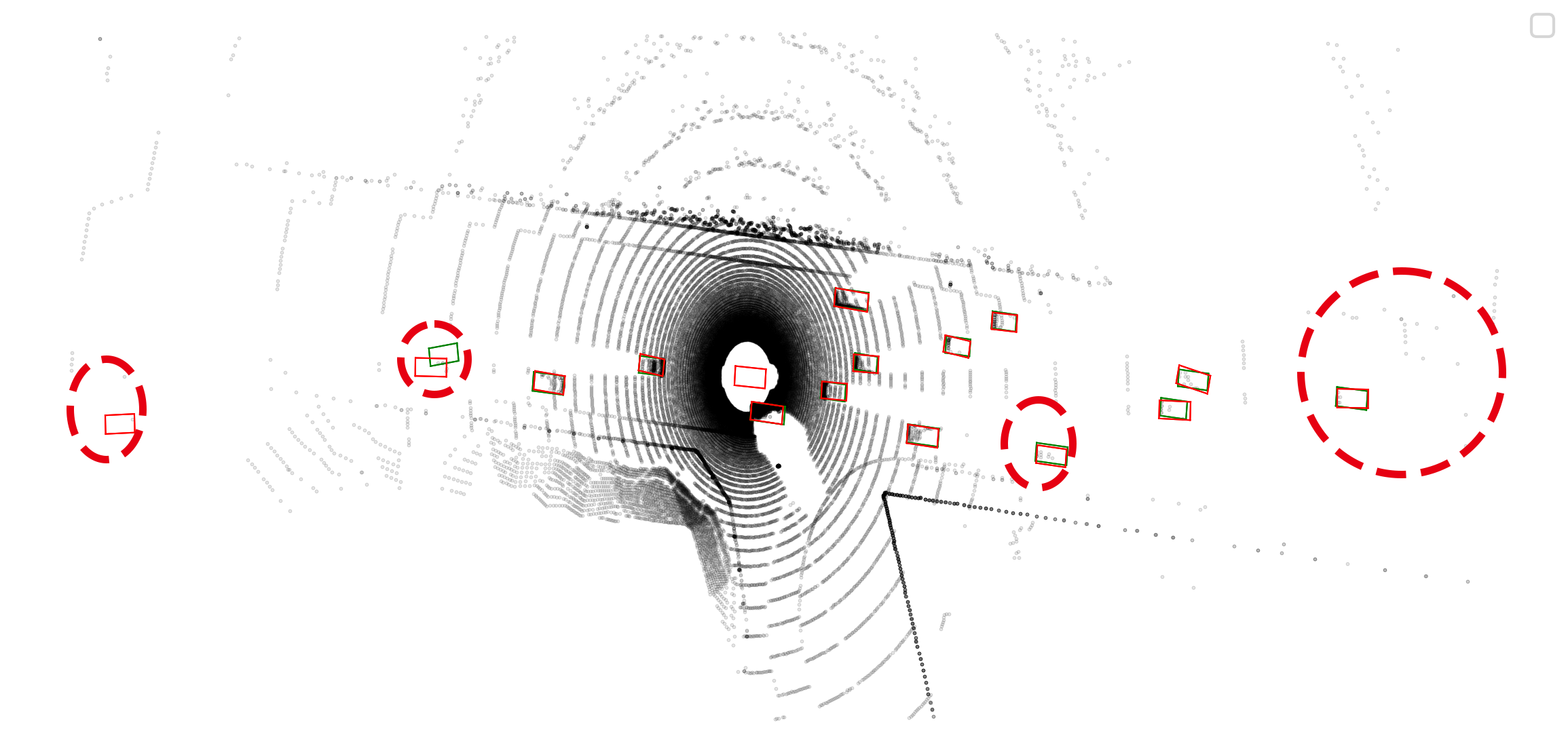}%
		\label{result_a}}
	\hspace{-6mm}
	\subfloat[]{\includegraphics[width=1.8in]{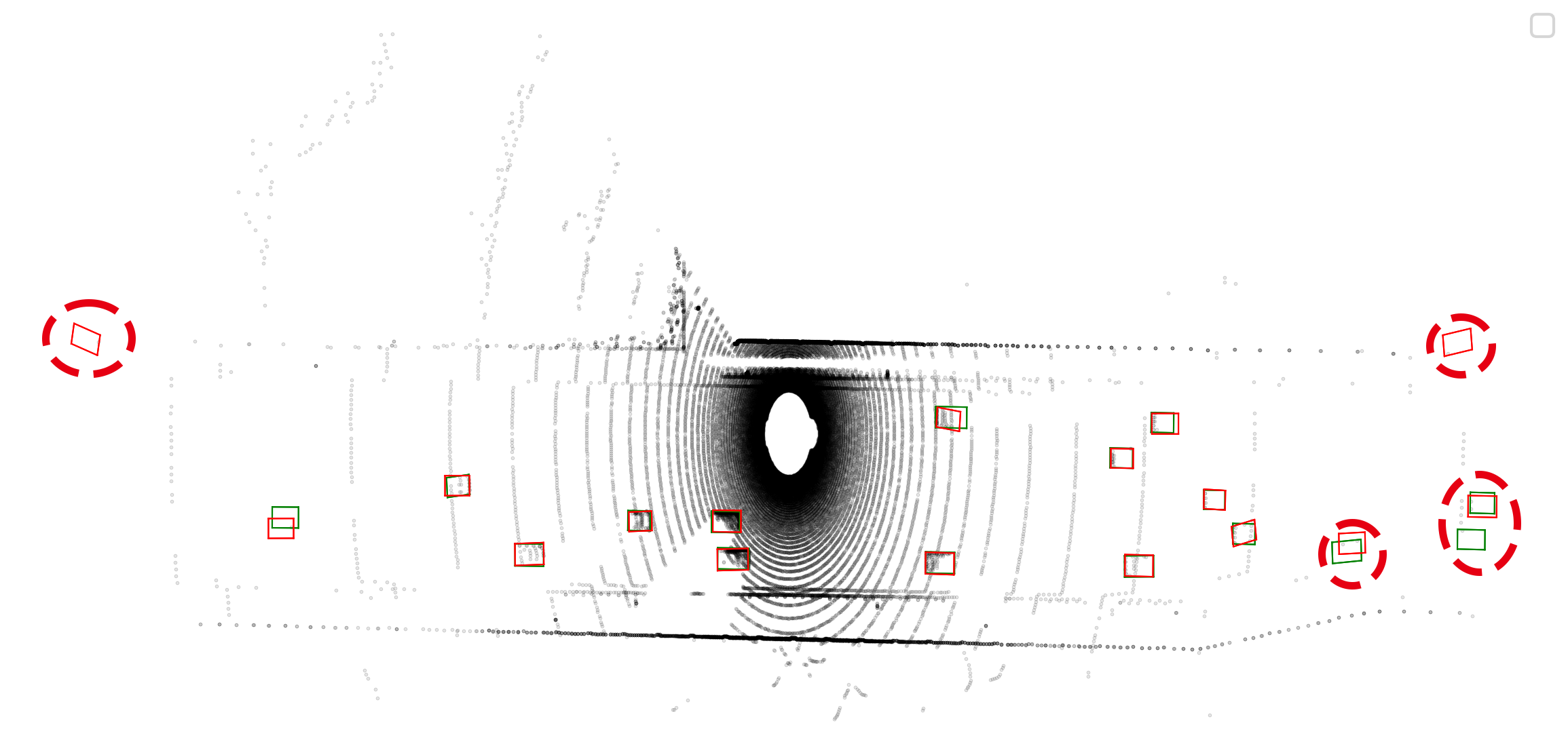} \includegraphics[width=1.8in]{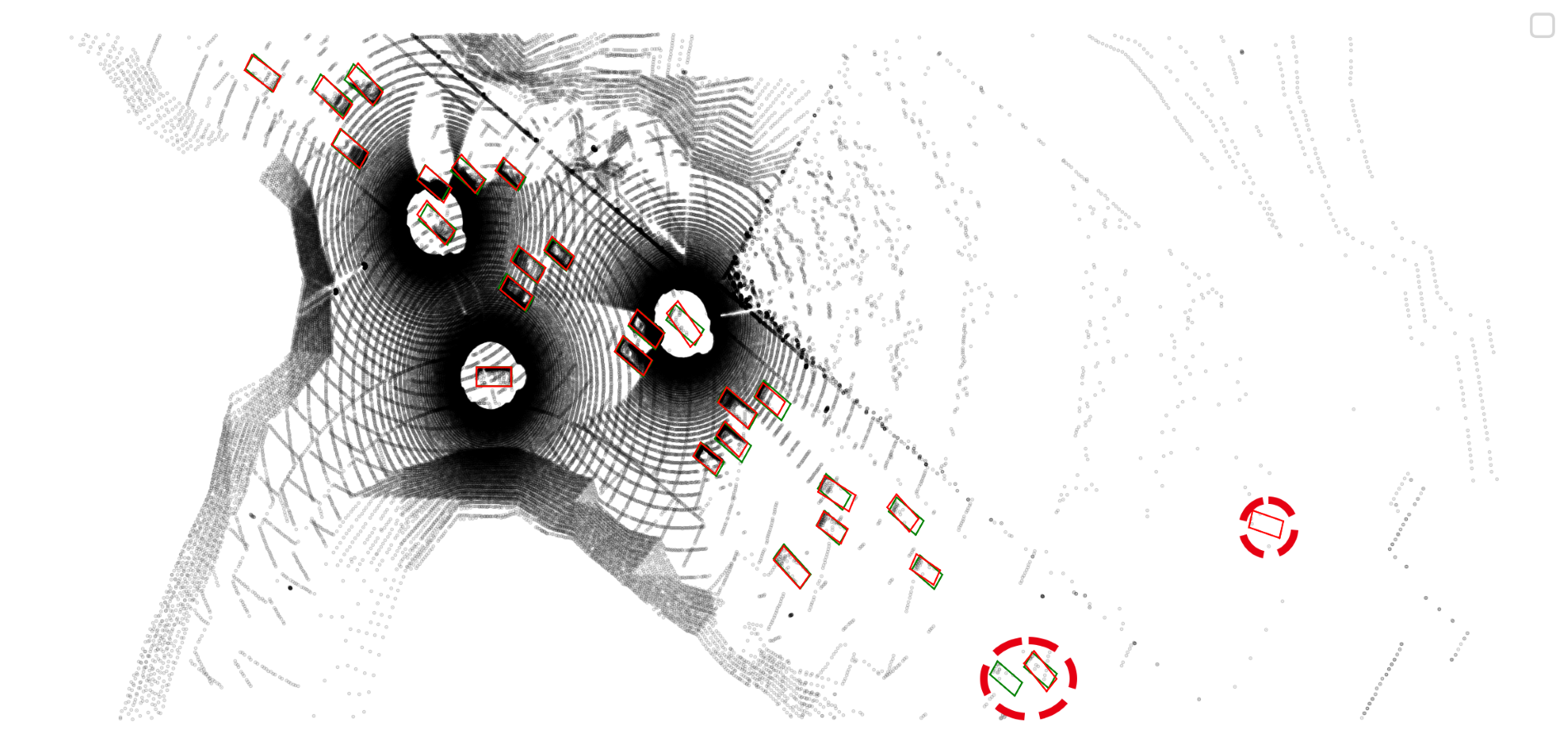}  \includegraphics[width=1.8in]{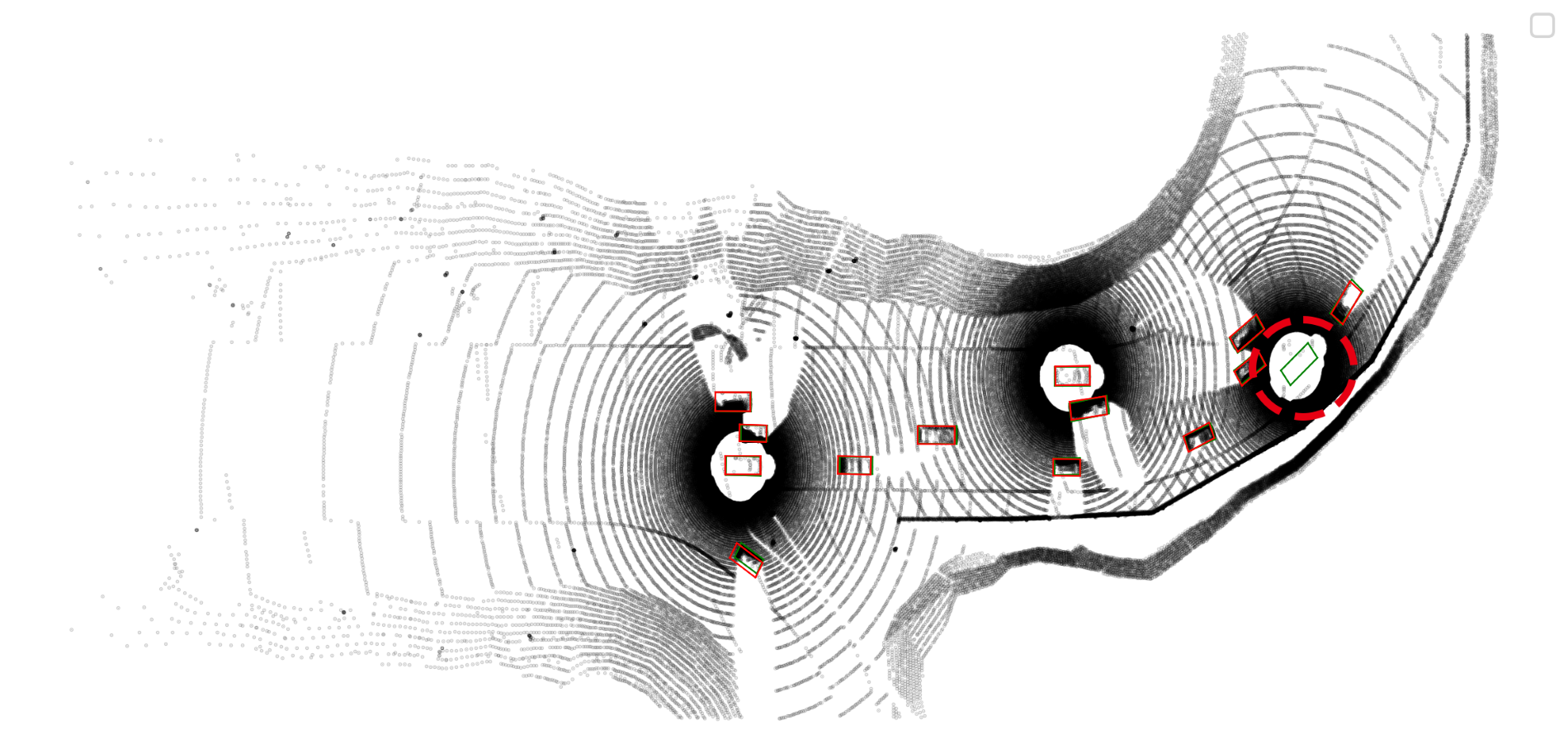} \includegraphics[width=1.8in]{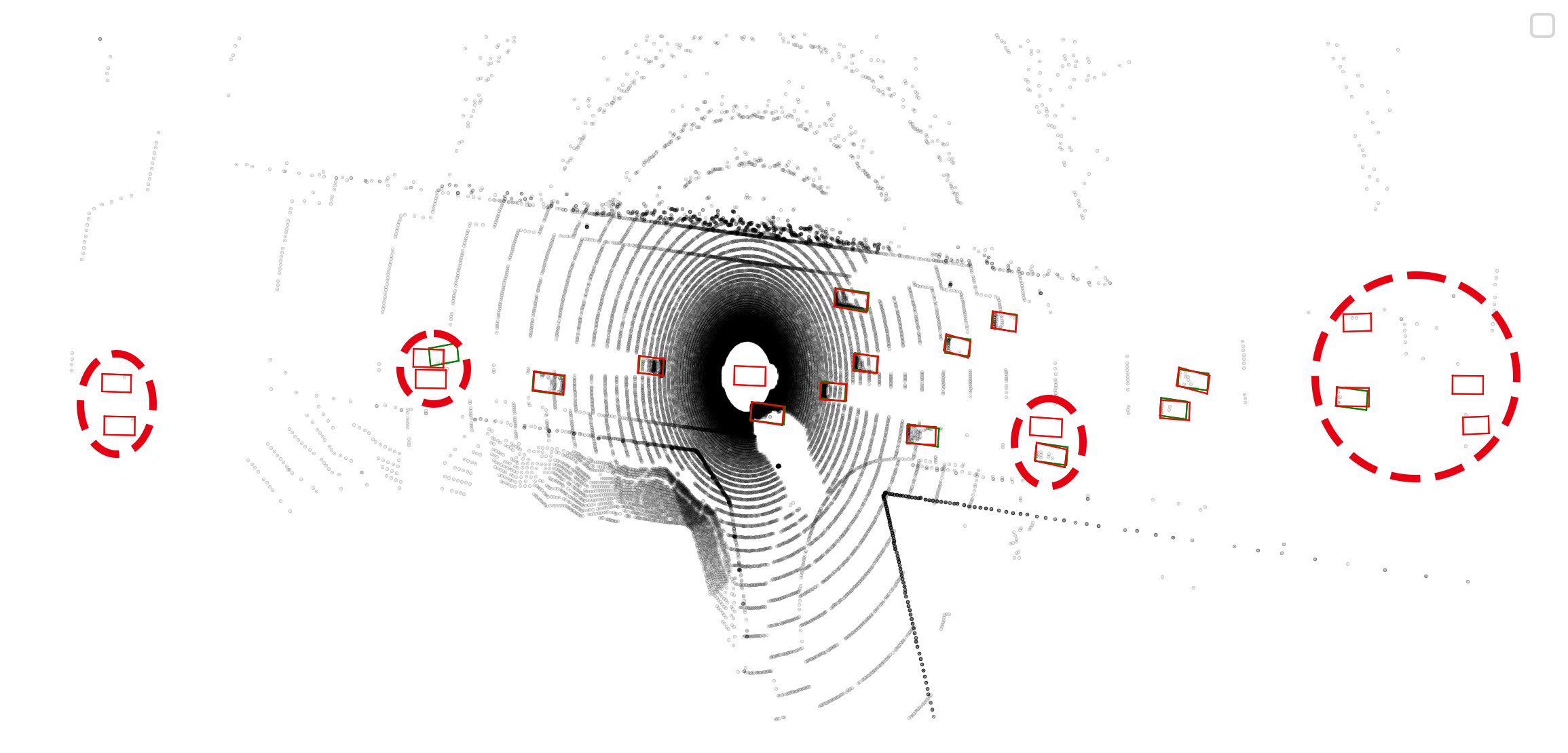}%
		\label{result_b}}
	\caption{Visualization of detection results. Green and red boxes represent ground-truth and detection results, respectively. Fig.~(a):~Detection results of \textit{Fast2comm}. Fig.~(b):~Detection results of baseline. \textit{Fast2comm} achieves more accurate and robust detection results, with fewer false positives and missed detections.} 
	\label{fig:results}
\end{figure*}

Figure~\ref{fig:results} presents the detection results of the proposed method and the baseline on the OPV2V dataset. Compared to the baseline, \textit{Fast2comm} achieves more accurate and robust detection results, exhibiting fewer false positives and missed detections.
This improvement is primarily due to the fact that Where2Comm~\cite{hu2022where2comm} directly uses the confidence map for sharing, which may include redundant background information. In contrast, \textit{Fast2comm} generates more accurate confidence features through prior supervision and selects critical prior features for sharing, thereby reducing redundant information and enhancing robustness against localization errors.
\subsection{ablations}
To validate the effectiveness and synergy of the proposed method, we conducted three ablation experiments based on the baseline model. The experimental results are summarized in Table~\ref{tab:table2}.
When the \textit{Confidence Feature Generation} (CFG) or \textit{GT Bbox-Based Feature Selection} (GT-FS) module was introduced individually, the model performance decreased. For instance, on the OPV2V dataset, the baseline achieved 78.44\% AP@0.7, while baseline+CFG dropped to 77.6\%, and baseline+GT-FS further decreased to 71.82\%. This phenomenon suggests that when CFG is applied alone, the additional supervision signal may not fully align with the original detection targets, introducing disturbances in the model's optimization direction. Although GT-FS incorporates spatial prior information, in the absence of prior supervised guidance, the generated feature maps may fail to accurately capture the target location, leading to redundant background information being shared.
It is noteworthy that when both CFG and GT-FS are introduced simultaneously, the model performance significantly surpasses that of the baseline. This indicates a strong synergistic effect between the two proposed modules: CFG enhances the discriminative capability of target features, while GT-FS effectively focuses on key spatial regions. Their combination greatly improves the model?s spatial representation ability and overall performance.
\begin{table}[!t]
	\caption{Ablation study results of the proposed core methods on datasets OPV2V and V2XSet. \textbf{CFG}:Confidence Feature Generation; \textbf{GT-FS}:GT Bounding Box-Based Feature Select\label{tab:table2}}
	\centering
	\begin{tabular}{c|c|c|c}
		\toprule[1pt]
		CFG & GT-FS & \begin{tabular}[c]{@{}c@{}}OPV2V \\ AP@0.5/0.7\end{tabular} & \begin{tabular}[c]{@{}c@{}}V2XSet \\ AP@0.5/0.7\end{tabular} \\
		\midrule
		&  & 87.80/78.44 & 82.04/68.73\\ 
		\checkmark &  & 87.95/77.62 & 83.71/68.17 \\
		& \checkmark & 86.41/71.82 & 79.10/52.32\\
		\checkmark & \checkmark & \textbf{88.86/79.62} & \textbf{84.71/71.61} \\
		\bottomrule[1pt]
	\end{tabular}
\end{table}
\section{Conclusion}
This paper introduces \textit{Fast2comm}, a communication-efficient and collaboration-robust multi-agent perception framework based on prior knowledge. The key innovation lies in generating foreground-background distinct confidence maps through prior supervision, followed by GT Bounding Box-based spatial prior feature selection to select and share only the most critical prior knowledge. Simultaneously, we decouple feature fusion into separate training and testing phases to optimize bandwidth utilization. Comprehensive experiments show that \textit{Fast2comm} achieves a trade-off between perception performance and communication bandwidth while maintaining superior robustness under varying localization errors.

\bibliographystyle{IEEEtran}
\bibliography{ref}

\end{document}